\documentclass[letterpaper, 10 pt, journal, twoside]{IEEEtran}  

\IEEEoverridecommandlockouts

\usepackage{float}
\usepackage{amsmath} 
\usepackage{amssymb}  
\usepackage[dvips]{graphicx}
\usepackage{graphicx}

\usepackage{caption}
\usepackage{subfig}
\usepackage{multirow}
\usepackage{amsfonts}
\usepackage{hyperref}
\usepackage{tabularx}
\usepackage{algorithmic}
\usepackage{algorithm}
\usepackage{url}
\usepackage{bm}
\usepackage{breakurl}
\usepackage{enumerate}
\usepackage[usenames,dvipsnames]{color}

\title{\LARGE \bf
Personalized Human-Swarm Interaction through Hand Motion
}

\author{
Matteo Macchini$^*$, \textit{Student Member, IEEE}, Ludovic De Matteïs$^*+$, Fabrizio Schiano$^*$, \textit{Member, IEEE}, \\ and Dario Floreano$^*$, \textit{Senior Member, IEEE}
\thanks{$^*$Laboratory of Intelligent Systems, École Polytechnique Fédérale de Lausanne, CH-1015 Lausanne (EPFL), Switzerland.}
\thanks{$^+$Ecole Normale Supérieure Paris-Saclay, Université Paris-Saclay, 91190 Gif-sur-Yvette, France.}%
}

\newcommand{\kron}{\otimes}

\newcommand{\matr}[1]{\boldsymbol{#1}}

\newcommand{\zeros}[2]{
\ifthenelse{\equal{#2}{1}}{\vect{0}_{#1}}{\matr{\cancel{O}}_{#1 \times #2}}
}
\newcommand{\ones}[2]{
\ifthenelse{\equal{#2}{1}}{\vect{1}_{#1}}{\matr{1}_{#1 \kron #2}}
}

\newcounter{simulationcase}

\newcommand{\mm}[1]{\colorbox{NavyBlue}{\color{white}   \textsf{\textbf{mm}}} \textcolor{NavyBlue}{ #1}}

\newcommand\vect[1]{{\boldsymbol{#1}}}

\renewcommand{\mm}[1]{}  

\definecolor{mygray}{gray}{0.75} 
\newcommand{\old}[1]{}  

\newcommand\reffig[1]{Fig.\ref{#1}}

\begin{document}

\maketitle
\thispagestyle{empty}
\pagestyle{empty}

\begin{abstract}

The control of collective robotic systems, such as drone swarms, is often delegated to autonomous navigation algorithms due to their high dimensionality.
However, like other robotic entities, drone swarms can still benefit from being teleoperated by human operators, whose perception and decision-making capabilities are still out of the reach of autonomous systems. Drone swarm teleoperation is only at its dawn, and a standard human-swarm interface (HRI) is missing to date. In this study, we analyzed the spontaneous interaction strategies of naive users with a swarm of drones.  We implemented a machine-learning algorithm to define a personalized Body-Machine Interface (BoMI) based only on a short calibration procedure. During this procedure, the human operator is asked to move spontaneously as if they were in control of a simulated drone swarm.
We assessed that hands are the most commonly adopted body segment, and thus we chose a LEAP Motion controller to track them to let the users control the aerial drone swarm. This choice makes our interface portable since it does not rely on a centralized system for tracking the human body.
We validated our algorithm to define personalized HRIs for a set of participants in a realistic simulated environment, showing promising results in performance and user experience. 
Our method leaves unprecedented freedom to the user to choose between position and velocity control only based on their body motion preferences.
\end{abstract}

\section{Introduction}\label{sec:introduction}

Recent advancements in robotic swarms and collective behavior are opening to new perspectives in many fields, such as collective transportation, surveillance, and mapping~\cite{cortes2017coordinated,hunt2020checklist}.
However, despite the concrete advantages of human capabilities with respect to autonomous navigation algorithms, no standard interfaces exist to date for such systems, and letting humans intervene on their behavior is still a challenge.
Here, we study the spontaneous motion patterns arising for naive users for human-swarm interaction, and leverage this knowledge to design and validate a machine learning algorithm capable of generating personalized human-swarm interfaces based on a user's preferred motion strategy.

The term \textit{telerobotics} identifies the branch of robotics involving a human operator controlling a robot situated in a different environment \cite{niemeyer_telerobotics_2008}.
Telerobotics is needed in all the tasks for which robotic autonomy is still incapable of achieving a sufficient level of performance \cite{abbink_topology_2018}.
Relevant examples include, but are not limited to, search-and-rescue, exploration of challenging environments, minimally invasive surgery \cite{bodner_first_2004, diftler_robonaut_2011, murphy_search_2008}.
As the robots dedicated to these applications rise in complexity and performance, a complementary effort is required to implement interfaces that are at the same time effective and comfortable for most users \cite{casper_human-robot_2003}.
However, standard interfaces like remote controllers still fail to achieve this goal and require substantial time and effort to be proficiently mastered by inexperienced users \cite{chen_human_2007, peschel_humanmachine_2013}.

Modern Human-Computer Interfaces (HCIs) tend to leverage the innate control capabilities that humans can exert over their body motion to provide more effective teleoperation systems for both Virtual Reality (VR) applications and telerobotics \cite{casadio_body-machine_2012}.
BoMIs have shown the potential to be more effective than standard interfaces both in terms of performance and of user experience, measured as a combination of cognitive workload necessary to control the robot and user engagement \cite{toet_toward_2020}.

For few decades, drones have attracted much attention both from researchers and industrial players~\cite{floreano2015science}. These robots achieved stunning levels of autonomy~\cite{LoiannoRAL2017,kaufmann2020deep} and we are now witnessing an effort in scaling the autonomy of single-drone systems to the so-called aerial swarms~\cite{chung2018survey,tahir2019swarms,coppola2020survey}. Indeed, groups of drones could unlock applications that are too complex, time-consuming, or even impossible for a single drone (e.g., collective transportation~\cite{villa2019survey}). However, a drone swarm, like other robotic entities, can still benefit from the high-level teleoperation and decision-making of a human operator.  
Different solutions allowed the user to control a single robot through hands, torso, or full-body motion \cite{macchini_hand-worn_2020, rognon_flyjacket:_2018, sanna_kinect-based_2013}. 
Recent work is dealing with the implementation of novel paradigms for swarm control \cite{schiano_rigidity_maintenance_2017,schiano_dynamic_2018,schilling2019learning}. In particular, we believe that the use of BoMIs for the operation of drone swarms is only at its dawn.
Most of these interfaces, though, rely on high-level commands such as "take off", "go right/left", and are not sufficiently sensitive for accurate navigation. 
Moreover, the implementation of a single, generic interface does not allow individuals to control the drone using their preferred motion strategy.
Based on an individual calibration procedure, personalized interfaces have shown superior results to generic ones in terms of learning time and performance \cite{macchini_personalized_2020, khurshid_data-driven_2015}.
However, research in motion-based HRIs for collective systems is not as developed, and few solutions have been proposed for this topic to date \cite{tsykunov_swarmtouch:_2018,aggravi2020connectivity}.
These HRIs are fixed, and thus do not account for individual motor preferences. Moreover, they rely on fixed mapping functions and directly map the user's hand position into the drone swarm center, possibly limiting the available workspace.  On the other hand, controlling the swarm velocity extends the workspace but can be less intuitive for naive subjects.


In this study, we propose a motion-based HRI to control of a drone swarm, which allows the user to both define their preferred strategy and choose between position and velocity control based on a calibration procedure.
The calibration consists of a physical demonstration of the preferred motion patterns used to control each Degree of Freedom (DoF) of the swarm~(Fig. \ref{f:protocol1}).
Briefly, we first observed the spontaneous motion patterns of participants when interacting with the swarm. 
After reducing the sensor coverage to the user's hands, since we found it to be the most relevant body segment for the task, we run a second experiment to assess their motion variability.
As hand movements varied from subject to subject and correlated partly with the robots' positions and partly with their velocity, we decided to let this option free. Therefore, we extended our framework to define the control methodology (i.e., position or velocity) from the subject's motion demonstration \cite{macchini_personalized_2020}.
Finally, we evaluated our system in a teleoperation task in a simulation with a swarm with 4 drones.

\section{Pilot Study}\label{sec:pilots}

For the preliminary experimental stage, N=20 human subjects were recruited to participate in 2 experimental sessions.
Informed consent was obtained from everyone before the experiment and the study was conducted while adhering to standard ethical principles\footnote{The experiments were approved by the Human Research Ethics Committee of the École Polytechnique Fédérale de Lausanne.}.
The results of our studies consist of statistical analysis and observations of human factors in human-robot interaction activities, mainly the participant's body motion.
Due to the limited number of subjects per condition, we chose non-parametric methods to assess the significance of our results: we used the Kruskal-Wallis test to assess the equality of the medians of different groups \cite{kruskal_use_1952}.

\begin{figure}[h]
    \begin{center}
        \includegraphics[width=0.9\columnwidth]{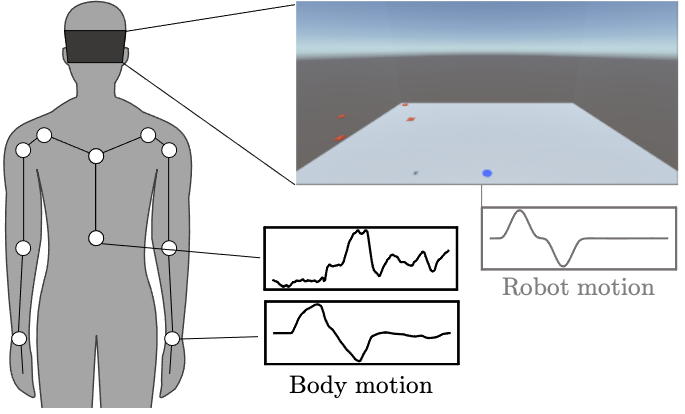}
        \caption{
        Experimental scenario for the identification of spontaneous motion patterns for aerial swarm teleoperation. The user sees a set of predefined actions performed autonomously by the robotic swarm in a simulated environment  (in the figure: left motion, the blue dot is used as reference). 
        The user moves as if they were controlling the motion of the drones and their body movements are tracked.}
        \label{f:protocol1}
    \end{center}
\end{figure}
\textbf{\textit{Acquisition of spontaneous body motion:}}
we implemented a swarm simulator, based on an existing quadrotor model\footnote{\href{https://github.com/UAVs-at-Berkeley/UnityDroneSim}{https://github.com/UAVs-at-Berkeley/UnityDroneSim}}. We configured the swarm to move according to a list of predefined maneuvers.
A master drone is controlled by the algorithm (or the operator), while the collective behavior of the slave agents is based on the Reynolds' algorithm~\cite{reynolds1987flocks,soria2019influence}. 
This potential-field based control algorithm consists of three rules: 1) a repulsive separation term to steer nearby drones away from each other, 2) a cohesion term to keep the drones close to each other, and 3) an alignment term that aligns the velocity vector of the agents. The respective acceleration terms for cohesion, separation, and alignment can be formalized as:
\begin{align*}
    \boldsymbol{a}_{i} &=  \boldsymbol{a}^{coh}_i + \boldsymbol{a}_{i}^{sep} + \boldsymbol{a}_i^{ali}\\
    \boldsymbol{a}_{i}^{coh} &= k^{coh}\frac{1}{|\mathcal{A}_i|}\sum_{j\in \mathcal{A}_i}\boldsymbol{r}_{ij}\\
    \boldsymbol{a}_{i}^{sep} &= k^{sep}\frac{1}{|\mathcal{A}_i|}\sum_{j\in \mathcal{A}_i}\frac{\boldsymbol{-r}_{ij}}{||\boldsymbol{r}_{ij}||}\\
    \boldsymbol{a}_{i}^{ali} &= k^{ali}\frac{1}{|\mathcal{A}_i|}\sum_{j\in \mathcal{A}_i} \boldsymbol{v}_i-\boldsymbol{v}_j
\end{align*}

where $\mathcal{A}_i$ represents the set of neighbors of the $i$-th agent, $\boldsymbol{r}_{ij}$ denotes the relative distance between of the $j$-th and the $i$-th agent, and $\boldsymbol{v}_{i}$ and $\boldsymbol{v}_{j}$ are the velocities of the $i$-th and the $j$-th agents, respectively.

The swarm was visualized in third-person view through a Head-Mounted Display (HMD) and performed 8 distinct maneuvers: right-left motion, up-down motion, front-back motion, and expansion-contraction, for a total of 4 DoF.
We showed the simulation to N=10 subjects and asked them to move as they were controlling the drone swarm with their body motion.
We used a motion capture system\footnote{\href{https://optitrack.com}{https://optitrack.com}} to track the kinematics of the user's upper body during the experiment (\reffig{f:protocol1}).
The human upper body was modeled as a kinematic chain consisting of 9 rigid bodies interconnected by spherical joints.
\\

\begin{figure}[t]
\begin{center}
  \includegraphics[width=1\columnwidth]{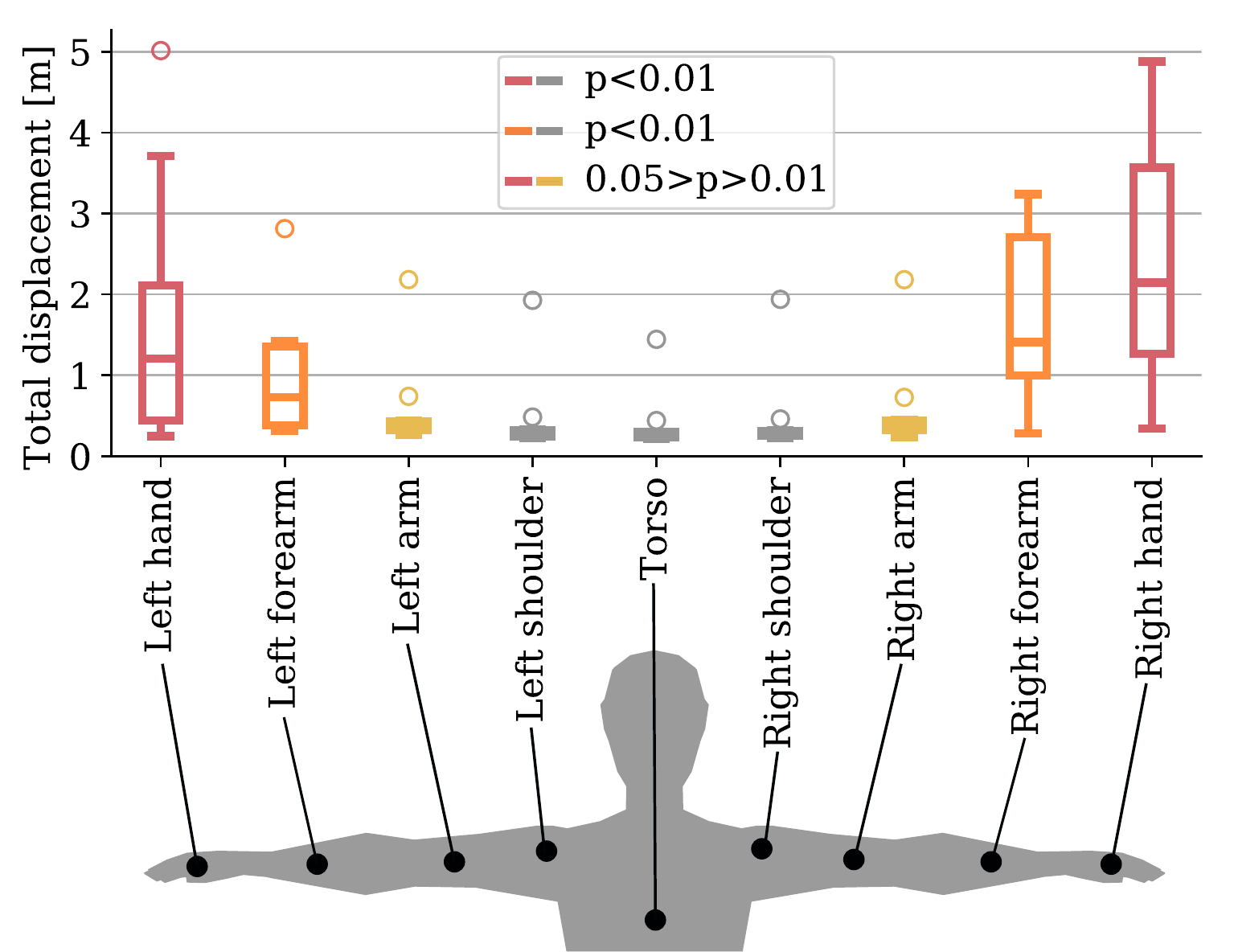}
\caption{
Results of the spontaneous motion acquisition experiment. The total motion of the center of mass of the different body segments shows a preponderant use of hands and forearms. The color code represents the statistical significance of equality of medians in the distributions.}

\label{f:motion}
\end{center}
\end{figure}

\textbf{\textit{Pilot Study 1 - Hand motion is the most relevant motion feature for the task:}}
we studied the participants' motion to identify the body segments which are most relevant for the task.
To do so, we analyzed the total displacement of the different body segments.
We found that higher displacement was associated with forearms and hands positions than with the rest of the upper body segments, demonstrating that users spontaneously prefer to use their hand motion to control the swarm (\reffig{f:motion}).
Specifically, most of the participants mimicked the motion of the drone using one or both hands, resulting in high displacement values for the right ($2.34 \pm 1.47m$) and left hand ($1.31 \pm 1.10m$).
Consequently, we observed that the forearms moved on average more than the rest of the body: both the right  ($1.67 \pm 1.03m$) and the left one ($0.68 \pm 0.41m$). The unbalance between right and left body segments is due to most of the participants being right-handed.
We discovered that the displacement of these 4 body segments was significantly higher than the motion of the torso and shoulders (all below $0.3m$ on average, $p<0.01$).
Also, hand motion was higher than upper arms motion, at a less significant level ($0.05>p>0.01$).

\textbf{\textit{Acquisition of hand motion:}}
according to this result, we decided to track only the hands of the users and we used as an acquisition system the LEAP Motion controller\footnote{\href{https://www.ultraleap.com/product/leap-motion-controller/}{https://www.ultraleap.com/product/leap-motion-controller/}}. With this choice we moved to a more portable solution and added finger tracking.
The LEAP Motion controller models the human hand as a kinematic chain composed of 21 rigid bodies, 1 for the palm and 4 for each finger, and provides their poses (positions and rotations). 
As we observed the participants mainly used hand motion, (\reffig{f:motion}), we decided to retain the positions of palms and fingertips relative to the palm, for a total of 6 rigid bodies per hand.
We computed the relative positions of the fingers as follows:
\begin{align*}
     \boldsymbol{p}_F' =  \boldsymbol{q}_P( \boldsymbol{p}_F -  \boldsymbol{p}_P)
\end{align*}
Where $ \boldsymbol{q}_P$ is the quaternion describing the rotation of the palm, $ \boldsymbol{p}_P$ the position of the palm, and $ \boldsymbol{p}_F$ the position of the finger's bone in the LEAP Motion's frame.
The original position $ \boldsymbol{p}_F$ is replaced by the new position $\boldsymbol{p}_F'$.
Additionally, we included a grasp factor to account for this common motion synergy. 
It was represented as:
\begin{align*}
    \boldsymbol{g} = \frac{\sum_{i=1}^5 \sum_{j=i+1}^{5} \boldsymbol{r}_{Fij}}{10}
\end{align*}
where $\boldsymbol{r}_{Fij}$ is the distance between the tip of the $i$-th and the $j$-th finger of the hand.
In total, we acquire 22 kinematic variables for each hand consisting of the 3D coordinates of palm and fingertips, the 3D rotation of the palm, and the grasp factor.
Fingertip rotations were removed from our variable set as they highly correlate to their positions.

\begin{figure*}[h!]
\begin{center}
\includegraphics[width=\textwidth]{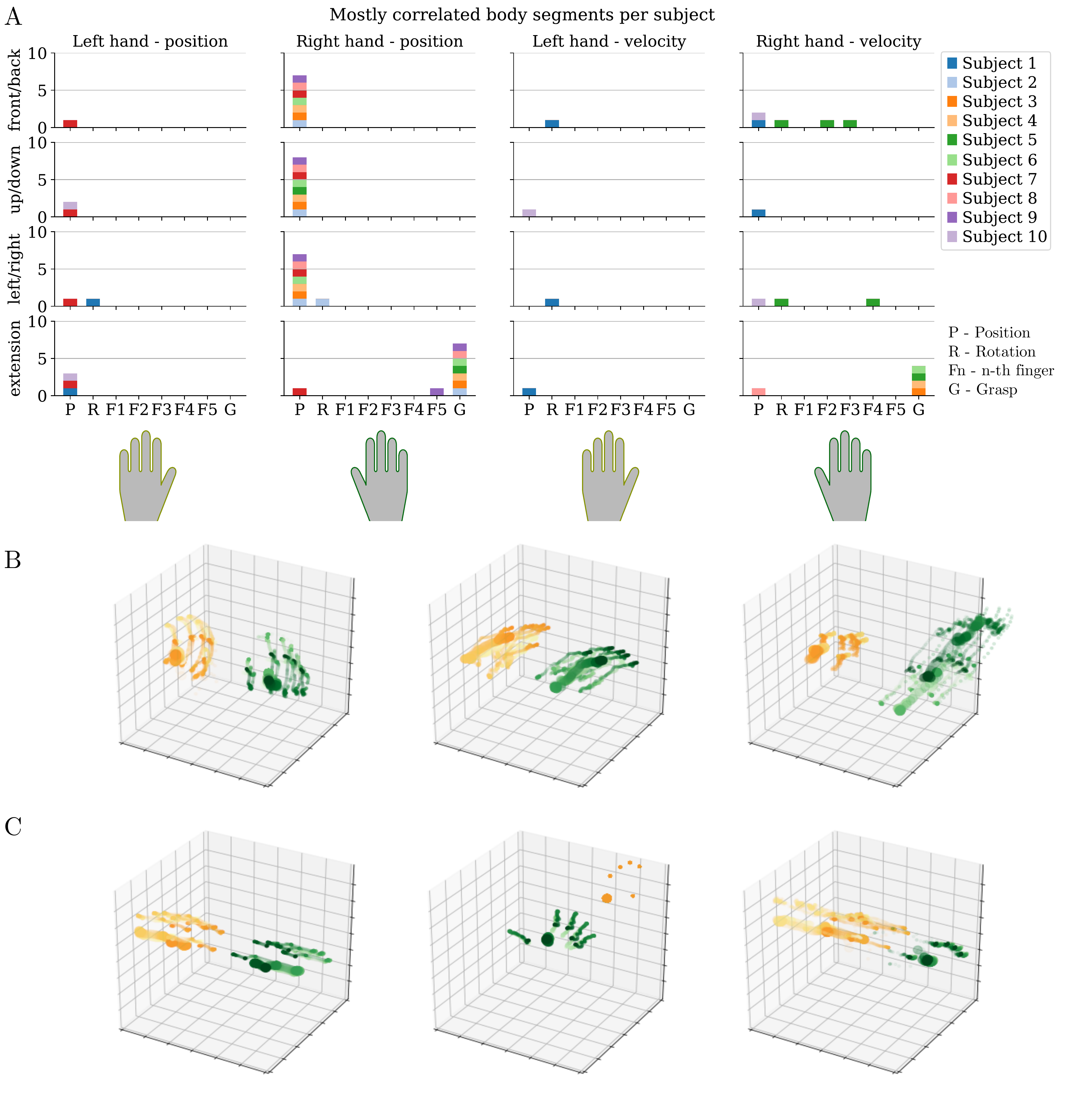}
	\caption{Variability of the observed motion strategies across experimental subjects. (A) Kinematic variables with the highest correlation between human and robot motion. Additional ones are added for a subject if their correlation score is at least $90\%$ of the first. Most subjects display individual motion traits, while only three of them (subj 3,4,5) show a full agreement. (B) Example of motion strategies for the different DoF of the swarm, front motion. From left to right, double-handed velocity control through palm rotation, double-handed position control through palm position, single-handed position control through palm position. (C) Example of motion strategies for the different DoF of the swarm, expansion of the swarm. From left to right, double-handed velocity control through palm proximity, single-handed position control through grasp synergy, asymmetric double-handed position control through palm proximity - asymmetric.}
	\label{f:variab}
\end{center}
\end{figure*}

\textbf{\textit{Pilot Study 2 -  Spontaneous hand motion is variable across users:}}
in presence of high agreement between the spontaneous motion of a population, it is possible to derive general interfaces that would fit all users.
We assessed the similarity of spontaneous hand motion among a set of users, to quantify the need for a personalized interface.
We asked 10 subjects to perform the same task as in the previous experiment, this time acquiring hand motion data from the LEAP Motion Controller.


We computed for each kinematic variable the Pearson's correlation coefficient $\alpha$ with respect to the swarm command, separately for each maneuver:
\begin{align*}
\alpha_{ij} = \frac{cov(X_i,Y_j)}{\sigma_{X_i}\sigma_{Y_j}}
\end{align*}
where $cov$ is the covariance of the $i$-th kinematic variable with the $j$-th robot maneuver, and $\sigma$ their standard deviation.
We normalized the correlation values on the features so that $\sum^i{\alpha_{ij}} = 1$.
In the case of velocity control, however, the participant's motion does not correlate with the swarm position (which we acquire as swarm action) but with its velocity.
For this reason, the Pearson's $\alpha$ coefficient was computed also between the swarm actions and the integral of the kinematic variables.

We found that, for each maneuver, the body segment mostly correlated with the robot motion was different for most subjects  (\reffig{f:variab}).
We retained body segments for which the correlation was at least $90\%$ of the maximally correlated one, showing a high variability.
Our data show that while some subjects spontaneously moved according to the position of the swarm, others tended to move their hands according to its velocity.

We identified 4 common motion patterns: right hand position control for the control of the 3D position of the drone, and right hand grasp for contraction/expansion.
In total, only $30\%$ of the participants agreed in using these patterns, while the others presented individual variants.
Specifically, for front/back motion, $70\%$ of the participants implemented a position control, $60\%$ using the position of the right hand and one using both hands, and $30\%$ a velocity control, $10\%$ using the position of the right hand and the others using the rotation of right or left hand.
We refer the reader to \reffig{f:variab}A for more details on the remaining DoFs.

Based on these results, we decided to implement personalized HRIs for each user and to allow them to choose with their motion between position and velocity control for each degree of freedom of the robot.

\section{Method}\label{sec:method}

The framework used to create the personalized HRIs is based on our previous work \cite{macchini2019personalized}.
The framework has been extended to accept inputs from the LEAP Motion controller and modified to allow both position and velocity control.
Here, we summarize the main algorithmic steps, pointing out the novelties introduced for this study.

\textbf{\textit{Data acquisition and preprocessing:}}
a first extension of our previous implementation is the acquisition of data from a new device and the definition of a new list of kinematic variables based on human hand biomechanics.
The simulation environment and the data acquired during the imitation phase are described in Sec. \ref{sec:pilots}.
Data are collected while the subject imitates the swarm motion with their hand movements and preprocessed to obtain the relative fingertip position and the grasp coefficient.
We compute the integral throughout the imitation phase and add it to the hand motion dataset.
The integral values are reset at the beginning of each new maneuver to prevent minor displacements from being accumulated over time and affect the dataset.
Since we consider a miscellaneous set of 3D coordinates and angular data, for which normalization is essential, all the motion variables were normalized to zero mean and unit variance.

\textbf{\textit{Feature selection:}}
in order to regularize the regression step, we reduce the dimensionality of the dataset by extracting the most informative kinematic variables.
We rank the variables based on a quality factor and select the most informative ones based on a threshold.
Let $X_i$ be the $i-th$ kinematic variable, and $Y_j$ be the $j-th$ degree of freedom of the robot.
The quality factor of $X_i$ with respect to $Y_j$ is given by:
\begin{align*}
    \lambda_{ij} = \alpha_{ij} \times SNR_i^k
\end{align*}
Where $\alpha_{ij}$ is the Pearson' correlation coefficient computed between $X_i$ and $Y_j$, and $SNR_i$ the signal-to-noise ratio of the kinematic variable $X_i$.
$k$ is a coefficient used to compensate the lower SNR associated with the integral terms, which are inherently low-passed. We set $k=2$.

We noticed that the number of variables selected changed substantially depending on the user's motion strategy. 
As this aspect made selecting an optimal threshold more challenging, we modified the algorithm pipeline to select a reduced set of variables for each of the robot's DoF (motion on the x, y, and z axes, expansion/contraction).
We normalized the $\lambda$ values to have $\sum_i \lambda_{ij} = 1$, and ranked the variables based on the associated $\lambda_{ij}$. 
Subsequently, we choose the $M$ first variables, so that $\sum_{1<i<M} \lambda_{ij} \geq \tau$, and set the threshold $\tau=0.7$.
This modification makes the feature selection process more interpretable and meaningful for human supervision and provides a lower-dimensional set of variables for each DOF, simplifying the regression step.

\textbf{\textit{Regression:}}
we finally train a linear model to define the mapping function between the user's motion and the robot's actions. We use ridge regression with BIC-optimized ridge parameter to maximize the model's fit to the data while preventing overfitting \cite{Ghadban_ridge, Neath_BIC}. 
This improvement, together with the previously described separation of the feature selection for each DOF, allowed us to remove the regularization step in the original algorithm based on CCA and simplify our pipeline. 

\section{Experimental Results}\label{sec:experiments}


\begin{figure}[t]
\begin{center}
  \includegraphics[width=\columnwidth]{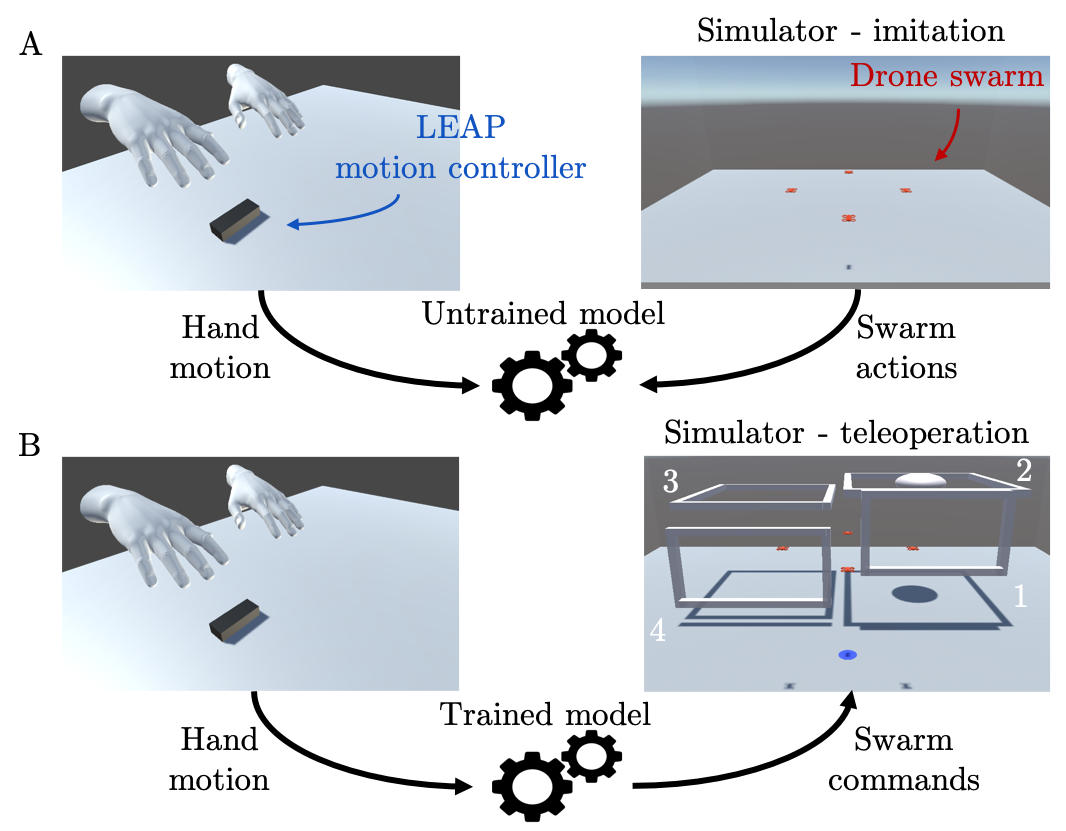}
\caption{
Experimental protocol for the qualification of the personalized HRI definition method. (A) Imitation task: the user's spontaneous interaction strategy is recorded and processed to implement a personalized interface. (B) Teleoperation task: the user controls the drone swarm through their personalized interface.
}
\label{f:protocol2}
\end{center}
\end{figure}
We recruited 10 participants to validate the effectiveness of the proposed method aimed at creating personalized HRIs for hand motion based swarm teleoperation.
The experiments consisted of 2 phases.
The first one was an imitation task, necessary for the user to show their preferred body motion to control the robot, and the second was the real teleoperation task (Fig. \ref{f:protocol2}).
In the first phase, we showed the same maneuvers described in Sec. \ref{sec:pilots} to the user through an HMD and asked them to move their hands accordingly (Fig. \ref{f:protocol2}A).
The teleoperation scenario consisted of a path composed of 4 gates (Fig. \ref{f:protocol2}B).
We instructed the participant to cross them in order from 1 to 4, performing the task as fast as possible while trying to avoid collisions.
The task was designed to require the robot swarm to be controlled in all its 4 DoF, as the gates are arranged in 3D on different altitudes and depth levels, and the spheric object inside gate 2 can be avoided only by expanding the swarm.
The participants were asked to steer the drone swarm across the path for a total of 10 times: 5 times through hand motion and 5 times using a standard remote controller (hereafter, conditions 'H' and 'R') to evaluate their performance prior and after training.
We pseudo-randomized the order of the interfaces to be used in order to compensate for the learning effects due to the user's increasing experience.

\begin{figure*}[t]
\begin{center}
\includegraphics[width=0.9\textwidth]{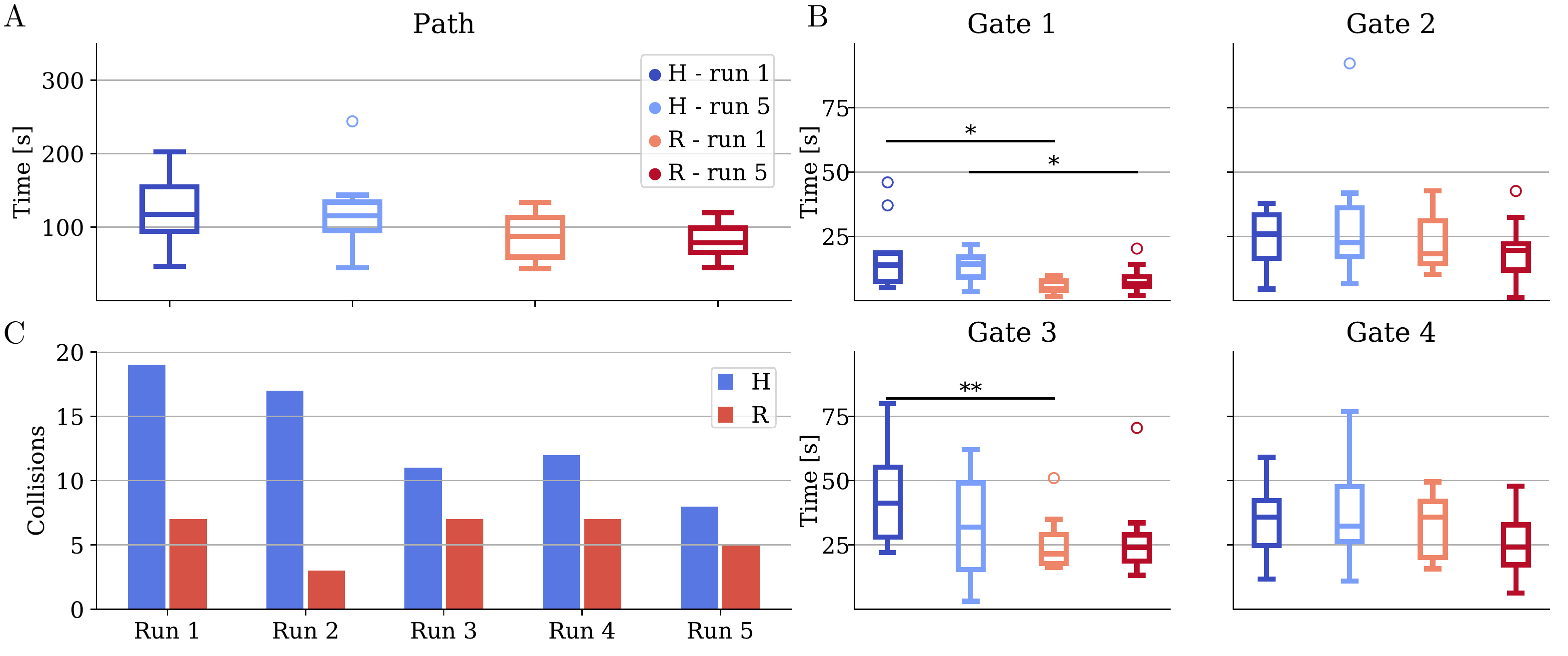}
	\caption{
Performance evaluation of groups using the remote controller (R) and the hand interface (H). (A) Total time needed to navigate along the path. While group R performed better before training, group H showed a higher learning rate. (B) Time needed to cross individual gates. Subjects showed significant task-dependent performance differences. (C) Number of occurred collisions during the navigation. Similarly to time performance, remote users performed better in the beginning, while hand interface users improved with training to reach comparable accuracy.}
	\label{f:res_alt}
\end{center}
\end{figure*}

\textbf{\textit{Remote controller users perform better initially, but hand interface users learn faster:}}
our results show that the teleoperation performance varies with both the used interface and the training for this task (Fig. \ref{f:res_alt}).
Group R outperformed group H in terms of time needed to navigate the whole path (Fig. \ref{f:res_alt}A). Similar results hold both in the first ($t^R_1 = 86.6 \pm 30.1s$, $t^H_1 = 120.1 \pm 44.5s$) and the last runs ($t^R_5 = 80.6 \pm 23.1s$, $t^H_5 = 105.1 \pm 30.1s$). However, group H showed a higher learning capability, reducing their time by $13.1\%$, in average, compared to the $7.0\%$ of group R.
Breaking down the path into the 4 inter-gate segments, however, we realized that the different maneuvers needed to steer the swarm through the gates were associated with different performance in our participants (Fig. \ref{f:res_alt}B).
Particularly, the time needed to cross gates 2 and 4 were similar for both groups, with a non-significant advantage for remote controller users.
Contrarily, group R performed significantly better in gate 3 before training ($t^R_1 = 22.4 \pm 6.1s $, $t^H_1 = 45.1 \pm 19.1s$, $p < 0.01$). The initial $100.1\%$ performance gap was reduced, with training, to $41.2\%$, non significant at a statistical level.
Finally, in crossing gate 1, group R performed closed to twice as fast both before ($t^R_1 = 5.46 \pm 2.71s $, $t^H_1 = 11.5 \pm 5.0s$, $p = 0.012$) and after training ($t^R_5 = 6.6 \pm 3.5s $, $t^H_5 = 13.1 \pm 5.3s$, $p = 0.011$).

We found that the use of different interfaces can affect the number of collisions during teleoperation (Fig. \ref{f:res_alt}C). 
Group R started with a lower number of collisions per run ($Coll^R_1 = 7$) than group H ($Coll^H_1 = 19$).
However, while Group R improved only marginally their performance ($Coll^R_5 = 5$), the hand interface users managed to reduce their collisions significantly with training ($Coll^H_5 = 8$).

\begin{figure}[h]
\begin{center}
  \includegraphics[width=\columnwidth]{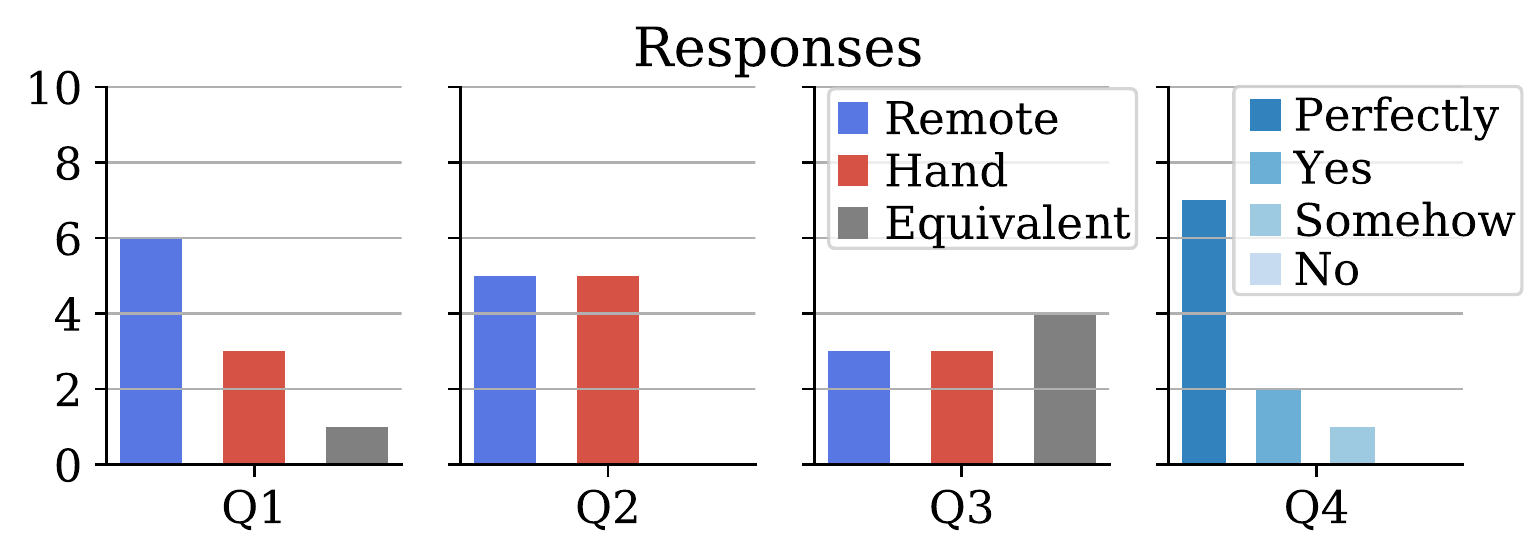}
\caption{
Survey results for questions 1-4. Despite the performance differences, users' preferences were split equally between the two interfaces, with a slight preference towards the hand interface to control the swarm position.
7/10 participants reported that the HRI was perfectly reflecting their desired behavior.
}
\label{f:quest}
\end{center}
\end{figure}

\textbf{\textit{User preference is equally split between interfaces:}}
after the teleoperation task, we asked our participants to fill a subjective feedback survey (Tab. \ref{t:quest}).
The questionnaire consisted of 4 multiple choice questions and 1 final feedback open question.
We asked two different questions about the control of the position and the expansion, as the second is a peculiar DoF of swarms and cannot be controlled with a single agent.
The responses to the survey show that the users did not find any of the two interfaces clearly superior (Fig. \ref{f:quest}).
In particular, the results were identical for the expansion/contraction DoF: 5 participants preferred the remote and 5 preferred the hand interface.
We obtained similar results for the general preference: 3 participants preferred the remote, 3 preferred the hand interface, and 4 did not have a preference.
Finally, 3 participants preferred the hand interface to control the position of the swarm, and only 1 preferred the remote.
7 participants declared that the interface reflected perfectly their expected motion, 2 that it was adequately accurate, and only 1 that his motion was somehow reflected in the interface.

In the final open question, 5 subjects mentioned that they felt a faster improvement when using the hand interface with respect to the remote. 
4 subjects remarked that their prior experience in using a remote controller might be the reason for their higher performance in the initial trial.
4 participants stated that they found the motion-based interface more engaging than the standard solution (specifically: "funny", "attractive", "impressive").
Finally, 3 participants responded that the limited field of view of the sensor affected their performance during the task when using the hand interface.

\begin{table}[h]
\caption{Personal feedback questionnaire}
\normalsize
\begin{center}
\begin{tabular}{  c p{7cm} } 
 \textbf{ID} & \textbf{Question} \\ 
 \hline
 Q1 & Which interface did you prefer to control the position? \\ 
 Q2 & Which interface did you prefer to control the expansion? \\ 
 Q3 & Which interface did you prefer in general? \\ 
 Q4 & Did the interface match the one you imagined during the calibration phase? \\
 Q5 & Please give your personal feedback on the teleoperation experience \\
  \hline
\end{tabular}
\label{t:quest}
\end{center}
\end{table}

\section{Discussion}\label{sec:conclusion}

This study investigates the motion strategies arising from the spontaneous interaction between a human and an aerial drone swarm and propose a solution to automatically derive a human-swarm interface from a user's spontaneous motion data.
We ran two pilot studies (N=20) to identify the most relevant body segments for this interaction and their motion characteristics.
Based on the results from this phase, we designed a machine learning-based algorithm to map a user's spontaneous motion into commands for the swarm and characterized it in a final user study (N=10). 
Here, we discuss our main findings.

Our pilot studies produced two major results.
First, by observing the full-body motion of a set of participants imitating the swarm behavior, we realized that the most engaged body segment was the users' hands.
On average, our participants moved their right hand $920\%$ and their left hand $515\%$ more than the torso, reflecting the statistical predominance of right-handedness in the population (Fig. \ref{f:motion}).
This result is particularly relevant in the field of motion-based HRIs, as it contrasts with findings that emerged in prior studies on different robots. 
When asked to mimic the flight of a fixed-wing drone, subjects tend to imitate its roll and pitch angles using their torso, with a minor engagement of their hands \cite{macchini_personalized_2020}. In our case of teleoperation of a drone swarm, we observed the opposite. 
We suggest two possible explanations for this fact. First, the interaction with a different robot might affect a user's spontaneous motion patterns. Also, the higher complexity of a drone swarm in terms of controllable DoFs might induce the necessity to use more articulated body segments, such as the arms. Second, switching the viewpoint impact on a user's spontaneous motion in VR imitation tasks \cite{macchini_impact_2021}.
Ground-view viewpoints, particularly, are associated with a higher inter-subject motion variability.
However, we chose this viewpoint since it is the easiest to adopt in a drone swarm teleoperation scenario. The alternatives would be to either switch from an onboard camera to another single robot or try to give the operator a first-person view from multiple cameras, which might overload the operator's senses. 

The second finding related to our pilot study concerns the variability of the users' hand motion.
By observing the human-robot motion correlation, we found that only $30\%$ of our participants exhibited the same motion patterns to control all the 4 DoF of the swarm (Fig. \ref{f:variab}).
This poor agreement is related to the hand segments interested in the motion patterns and the type of controller that the participants imagined for the drone swarm. 
While some users moved their hand as if they were controlling the position of the swarm (for instance, placing their hand in the same position as the center of the virtual agents), others moved as if they were controlling its velocity. We proved this by correlating the hands' kinematic variables and their integrals with the swarm actions.

Based on these results, we decided to define personalized mapping functions for each subject.
We implemented a set of modifications to an existing algorithm to allow the use of the new sensor (the LEAP Motion controller), and to allow the user to choose the control mode (position/velocity) through their motion.
The results relative to the teleoperation user study provide new insights in motion-based teleoperation of robotics swarms. First, we found that the use of remote controllers leads to a shorter time needed to complete the task prior to training (Fig. \ref{f:res_alt}A). 
Due to the widespread popularity of remote controllers for several applications, from gaming to teleoperation, it is today nearly impossible to recruit participants who are naive to their use.
However, group H improved their performance almost twice as fast as group R. This result might be due to the lower initial performance, and was perceived by our participants and reported as subjective feedback in our survey.
We observed similar results regarding the number of collisions that occurred during the task (Fig. \ref{f:res_alt}B).
While group R never had more than 7 accidents during the 5 repetitions of the task, group H managed to improve from 19 collisions in the first run to 8, after training ($-58\%$).
We also found that this effect was only true for some sections of the proposed path: user's time need to cross 2 out of 4 gates were similar with both interfaces, while significantly changed for the remaining two gates (Fig. \ref{f:res_alt}B)
This result suggests that some maneuvers (like, in this case, crossing a horizontal and vertical gate) can be more challenging using body motion, while others are equally hard to perform with different interfaces.
Possible explanations of such effect could be related to the human's perturbed depth perception, and proprioceptive capabilities in virtual environments \cite{ingram_proprio,chen_human_2007}.
Finally, the subjective feedback survey demonstrated that users did not prefer one of the two interfaces and that the personalization was accurate.

Despite the lower performance for untrained users, our study provides encouraging results: all the participants were able to navigate the drone swarm through the path, using a partly position-based, partly velocity-based control paradigm adapted to their preferences. To our knowledge, this work provides the first algorithm capable of such a level of personalization. 
Moreover, the simplicity of the calibration procedure  allows a user to change their strategy in a matter of minutes, without redesigning the interface.
As one of our subjects observed, it is known that humans are not able to determine the easiest way to control robots based on their spontaneous motion \cite{macchini_does_2020}, and so a different control strategy might seem more effective in hindsight.
With our method, it is possible to generate a new interface in less than two minutes.

This work opens to new intriguing options for future research. 
First, it would be a valuable addition to test the transferability of the proposed approach on a real drone swarm and to study if our results hold in the real world. 
Moreover, the conception of a method to perform first-person view control of robotics swarms could extend this work to this common viewpoint for teleoperation.
Also, as we observed that a maneuver-related effect on the interface performance, it would be interesting to verify the consistency of performance with a set of different paths.
Finally, given the high variability, we observed in the motion of our subjects, increasing the participants pool to include a more varied population would undoubtedly add value to our work.

\section{Conclusions}

In this paper, we applied a methodology to identify the relevant motion patterns for human-swarm interactions through body motion and assessed their variability among different individuals. 
Our results showed that the hand body segment was the most commonly adopted one and that most people tend to use their hands in significantly different ways. We extended an HRI learning framework to allow users to control a drone swarm through their preferred hand strategy.
The proposed interfaces showed promising performance and provided a convincing user experience. The possibility to develop quickly new interfaces through an imitation task is a significant contribution, which could impact the future design of human-swarm interaction systems.


\section{Acknowledgements}

This work was partially funded by the European Union’s Horizon 2020 research and innovation  programme under grant agreement ID: 871479  AERIAL-CORE, the Swiss National Science Foundation (SNSF) with grant number 200021-155907, and the National Centre of Competence in Research (NCCR) Robotics.








\bibliographystyle{IEEEtran}
\bibliography{IEEE_journals_full.bib,IEEE_conferences_full.bib,references}

\end{document}